# The Reaction RuleML Classification of the Event / Action / State Processing and Reasoning Space

Adrian Paschke

————   http://ibis.in.tum.de/research/ReactionRuleML/classification.htm————

**Abstract**— Reaction RuleML is a general, practical, compact and user-friendly XML-serialized language for the family of reaction rules.

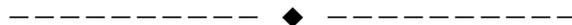

# Concepts and History in Event / Action / State Processing

In this white paper we give a review of the history of event / action /state processing and reaction rule approaches and systems in different domains, define basic concepts and give a classification of the event, action, state processing and reasoning space as well as a discussion of relevant / related work

## 1 Overview on Event / Action Processing

**Active Databases and ECA Rule Systems**

Active databases are an important research topic due to the fact that they find many applications in real world systems and many commercial databases systems have been extended to allow the user to express **active rules** whose execution can update the database and trigger the further execution of active rules leading to a cascading sequence of updates (often modelled in terms of execution programs). Several active database systems have been developed, e.g. ACCOOD [1], Chimera [2], ADL [3], COMPOSE [4], NAOS [5], HiPac [6]. These systems mainly treat event detection and processing purely procedural and often focus on specific aspects. In this spirit of procedural ECA formalisms are also systems such as AMIT [7], RuleCore [8] or JEDI [9]. Several papers discuss formal aspects of active databases on a more general level – see e.g. [10] for an overview. Several event algebras have been developed, e.g. Snoop [11], SAMOS [12], ODE [4]: The object database ODE [4] implements event-detection mechanism using finite state automata. SAMOS [12] combines active and object-oriented features in a single framework using colored Petri nets. Associated with primitive event types are a number of parameter-value pairs by which events of that kind are detected. SAMOS does not allow simultaneous atomic events. Snoop [11] is an event specification language which defines different restriction policies that can be applied to the operators of the algebra. Complex events are strictly ordered and cannot occur simultaneously. The detection mechanism is based on trees corresponding to the event expressions, where primitive event occurrences are inserted at the leaves and propagated upwards in the trees as they cause more complex events to occur. There has been a lot of research and development concerning knowledge updates in active rules (execution models) in the area of active databases and several techniques based on syntactic (e.g. triggering graphs [13] or activation graphs [14]) and semantics analysis (e.g. [15], [16]) of rules have been proposed to ensure termination of active rules (no cycles between rules) and confluence of update programs (always one unique minimal outcome). The combination of deductive and active rules has been also investigated in different approaches manly based on the simulation of active rules by means of deductive rules [17-19].

**Production Rule Systems and Update Rule Programs**

The treatment of active rules in active databases is to some extend similar to the **forward-chaining production rules system paradigm** in artificial intelligence (AI) research [20]. In fact, triggers, active rules and integrity constraints, which are common in active DBMS, are often implemented in a similar fashion to forward-chaining production rules where the changes in the conditions due to update actions such as "*assert*" or "*retract*" on the internal database are considered as implicit events leading to further update actions and to a sequence of "firing" production rules, i.e.: "*if Condition then Action*". There are many forward-chaining implementations in the area of deductive databases and many well-known forward-reasoning engines for production rules such as ILOG's commercial jRules system, Fair Isaac/Blaze Advisor, CA Aion, Haley, ESI Logist or popular open source solutions such as OPS5, CLIPS or Jess which are based on the RETE algorithm. In a nutshell, this algorithm keeps the derivation structure in memory and propagates changes in the fact and rule base. This algorithm can be very effective, e.g. if you just want to find out what new facts are true or when you have a small set of initial facts and when there tend to be lots of different rules which allow you to draw the same conclusion. This might be also on reason why production rules have become very popular as a widely used technique to implement large expert systems in the 1980s for diverse domains such as troubleshooting in telecommunication networks or computer configuration systems. Classical production rule systems and most database implementations of production rules [21-23] typically have an operational or execution semantics defined for them, but lack a precise theoretical foundation and do not have a formal semantics. Although production rules might simulate derivation rules via asserting a conclusion as consequence of the proved condition, i.e. "*if Condition then assert Conclusion*", the classical production rule languages such as OPS5 are less expressive since they lack a clear declarative semantics, suffer from termination and confluence problems of their execution sequences and typically do not support expressive non-monotonic features such as classical or negation-as-finite failure or preferences, which makes it sometimes hard to express certain real life problems in a natural and simple way. However, several extensions to this core production systems paradigm have been made which introduce e.g. negations (classical and negation-as-finite failure) [24] and provide declarative semantics for certain subclasses of production rules systems such as stratified production rules. It has been shown that such stratified production systems have a declarative semantics defined via their corresponding logic program (LP) into which they can be transformed [25] and that the well-founded, stable or preferred semantics for production rule systems coincide in the class of stratified production systems [24]. Stratification can be implemented on top of classical production rules in from of priority assignments between rules or by means of transformations into the corresponding classical ones. The strict definition of stratification for production rule systems has been further relaxed in [26] which defines an execution semantics for update rule programs based upon a monotonic fixpoint operator and a declarative semantics via transformation of the update program into a normal LP with stable model semantics. Closely related are also logical update languages such as transaction logics and in particular serial Horn programs, where the serial Horn rule body is a sequential execution of actions in combination with standard Horn pre-/post conditions. [27] These serial rules can be processed top-down or bottom-up and hence are closely related to the production rules style of "*condition* → *update action*". Several approaches in the active database domain also draw on transformations of active rules into LP derivation rules, in order to exploit the formal declarative semantics of LPs to overcome confluence and termination problems of active rule execution sequences. [28-30]. The combination of deductive and active rules has been also investigated in different approaches mainly based on the simulation of active rules by means of deductive rules. [17, 18] Moreover, there are approaches which directly build reactive rules on top of LP derivation rules such as the Event-Condition-Action Logic Programming language (ECA-LP) which enables a homogeneous representation of ECA rules and derivation rules [31].

**Event Notification Systems and Reaction Rule Interchange Languages**

Recently, a strong demand for event/action processing functionalities comes from the web community, in particular in the area of Semantic Web and Rule Markup and the upcoming W3c Rule Interchange Language (e.g. RuleML[1] or RIF[2]). In distributed environments such as the (Semantic) Web with independent agent / system nodes and open or closed domain boundaries event processing is often done using **event notification and communication mechanisms**. Systems either communicate events in terms of messages according to a predefined or negotiated communication/coordination protocol [32] and possibly using a particular language such as the FIPA Agent Communication Language (FIPA-ACL) or they subscribe to publishing event notification servers which actively distribute events (push) to the subscribed and listening agents using e.g. common format such as the common base event format [33]. Typically the interest here is in the particular event sequence which possibly follows a certain communication or coordination protocol, rather than in single event occurrences which trigger immediate reactions as in the active database trigger or ECA rules.

**Temporal KR Event / Action / Transition Logic Systems**

A fourth dimension to events and actions which has for the most part proceeded separately has the origin in the area of knowledge representation (KR) and logic programming (LP) with close relations to the formalisms of process and transition logics. Here the focus is on the development of axioms to formalize the notions of actions resp. events and causality, where events are characterized in terms of necessary and sufficient conditions for their occurrences and where events/actions have an effect on the actual knowledge states, i.e. they transit states into other states and initiate / terminate changeable properties called **fluents**. Instead of detecting the events as they occur as in the active database domain, the KR approach to events/actions focuses on the inferences that can be made from the fact that certain events are known to have occurred or are planned to happen in future. This has led to different views and terminologies on event/action definition and event processing in the **temporal event/action logics** domain. Reasoning about events, actions and change is a fundamental area of research in AI since the events/actions are pervasive aspects of the world in which agents operate enabling retrospective reasoning but also prospective planning. A huge number of formalisms for deductive but also abductive reasoning about events, actions and change have been developed. The common denominator to all this formalisms and systems is the notion of states a.k.a. fluents [34] which are changed or transit due to occurred or planned events/actions. Among them are the event calculus [35] and variants such as the interval-based Event Calculus (see section 3.3), the situation calculus [36, 37], features and fluents [34], various (temporal) action languages [38-42], fluent calculi [43, 44] and versatile event logics [45]. Most of these formalisms have been developed in relative isolation and the relationships between them have only been partially studied, e.g. between situation calculus and event calculus or temporal action logics (TAL) which has its origins in the features and fluents framework and the event calculus. Closely related and also based on the notion of (complex) events, actions and states with abstract models for state transitions and parallel execution processes are various process algebras like TCC [46], CSS [47] or CSP [48], (labelled) transition logics (LTL) and (action) computation tree logics (ACTL) [49, 50]. Related are also update languages [18, 31, 51-57] and transaction logics [27] which address updates of logic programs where the updates can be considered as actions which transit the initial program (knowledge state/base) to a new extended or reduced state hence leading to a sequence of evolved knowledge states. Many of these update languages also try to provide meaning to such dynamic logic programs (DLPs). However, unlike ECA languages and production rules these languages typically do not provide complex event / action processing features and exclude external calls with side effects via event notifications or procedural calls.

---
[1] http://www.ruleml.org
[2] W3C RIF

## 2 Basic Concepts in Event / Action Processing

**Definition 1:** *Atomic Events, Event Type Pattern Definitions and Event Context*

A raw event (a.k.a. atomic or primitive event) is defined as an instantaneous (occurs in a specific point in time), significant (relates to a context), atomic occurrence (it cannot be further dismantled and happens completely or not at all):

occurs(e,t).   event e occurs at time point t

We distinguish between event instances (simply called events) which occur and event type pattern definitions. An event type pattern definition (or simply event definition or event type) describes the structure of an (atomic or complex) event, i.e. it describes its detection condition(s). A concrete "instantiation" of a type pattern is a specific event instance, which is derived (detected) from the detection conditions defined within the rules body. This can be, e.g., a particular fact becoming true, an external event in an monitored system, a communication event within a conversation, a state transition from one state to another such as knowledge updates, transactions (e.g., begin, commit, abort, rollback), temporal events (e.g., at 6 p.m., every 10 minutes), etc.

Typically, events occur in a context that is often relevant to the execution of the other parts of the ECA rules, i.e., event processing is done within a context. A context can have different characteristics such as:

- Temporal characteristic designates information with a temporal perspective, e.g., service availability within one month or within 60 minutes from X.
- Spatial characteristic designates information with a location perspective, e.g. message reached end-point.
- State characteristic designates information with a state perspective, e.g., low average response time.
- Semantic characteristic designates information about a specific object or entity, e.g. persons that belong to the same role or messages that belongs to same conversation.

To capture the local context of an event, we use variables, i.e., the context information is bound to variables which can be reused in the subsequent parts of an ECA rule, e.g., in the action part.

**Definition 2** *Complex Events and Event Algebra*

A complex event type (i.e. the detection condition of a complex event) is built from occurred atomic or other complex event instances according to the operators of an event algebra. The included events are called components while the resulting complex event is the parent event. The first event instance contributing to the detection of a complex event is called initiator, where the last is the terminator; all others are called interiors.

**Definition 3** *Event Processing*

Event processing describes the process of selecting and composing complex events from raw events (event derivation), situation detection (detecting transitions in the universe that requires reaction either "reactive" or "proactive") and triggering actions as a consequence of the detected situation (complex event + conditional context). Examples are: "*If more than three outages occur then alert*" or "*If a department D is retracted from the database than retract all associated employees E*". Events can be processed in real time without persistence (short-term) or processed in retrospect as a computation of persistent earlier events (long-term), but also in aggregated from, i.e., new (raw) events are directly added to the aggregation which is persistent. According to the ECA paradigm event processing can be conditional, i.e. certain conditions must hold before an action is triggered. Event processing can be done either actively (pull-model), i.e. based on actively monitoring the environment and, upon the detection of an event, trigger a reaction, or passively (push-model), i.e., the event occurrences are detected by an external component and pushed to the event system for further processing.

Most event algebras in the active database domain treat (complex) events as instantaneous occurrences at a particular point in time, e.g. *(A;B)* in Snoop [11] is associated with the occurrence time of *B* (the terminator), i.e. the complex event is detected at the time point when *B* occurs. We have shown that this results in unintended semantics for some compositions in these algebras. [58] We further have demonstrate that these problems do not arise, if we associate the detection of a complex event with the maximum validity interval (MVI) in which it occurs, i.e., the interval in which all component events occurred, rather then the time of detection given by the terminator timestamp. This interval-based treatment of event definitions is a widespread assumption in the KR community and typical event logics such as the Event Calculus [35] enable temporal reasoning to derive the effects of events which hold at an interval beginning with an "initiating" event and ending with a "terminating" event. We extend the Event Calculus formalization of temporal event based reasoning with the notion of event intervals which hold between time intervals *[t1,t2]*. We use this new interval-based EC variant to redefine typical event algebra operators and investigate how far it is possible to reconcile the active database approach based on events detectable at a single point in time with the durative, inference-based treatment of events, occurring within an interval, in the sense of KR event/action logics. [58]

## 3 A Classification of the Event / Action / Processing Space [58]

### I) Classification of the Event Space

#### 1. Processing
(situation detection or event/action computation / reasoning)

- **Short term**: Transient, non-persistent, real-time selection and consumption (e.g. triggers, ECA rules): *immediate reaction*
- **Long term**: Transient, persistent events, typically processed in retrospective e.g. via KR event reasoning or event algebra computations on event sequence history; but also prospective planning / proactive, e.g. KR abductive planning: *defered or retrospective/prospective*
- **Complex event processing**: computation of complex events from event sequence histories of previously detected raw or other computed complex event (event selection and possible consumption) or transitions (e.g. dynamic LPs or state machines); typically by means of event algebra operators (event definition) (e.g. ECA rules and active rules, i.e. sequences of rules which trigger other rules via knowledge/state updates leading to knowledge state transitions)
- **Deterministic vs. non-deterministic**: simultaneous occurred events give rise to only one model or two or more models
- **Active vs. Passive**: actively detect / compute / reason event (e.g. via monitoring, sensing akin to periodic pull model or on-demand retrieve queries) vs. passively listen / wait for incoming events or internal changes (akin to push models e.g. publish-subscribe):

#### 2. Type

- **Flat vs. semi-structured compound data structure/type**, e.g. simple String representations or complex objects with or without attributes, functions and variables
- **Primitive vs. complex**, e.g. atomic, raw event or complex derived/computed event
- **Temporal**: Absolute (e.g. calendar dates, clock times), relative/delayed (e.g. 5 minutes after …), durable (occurs over an interval), durable with continuous, gradual change (e.g. clocks, countdowns, flows)
- **State or Situation**: flow oriented event (e.g. "server started", "fire alarm stopped")
- **Spatio / Location**: durable with continuous, gradual change (approaching an object, e.g. 5 meters before wall, "bottle half empty" )
- **Knowledge Producing**: changes agents knowledge belief and not the state of the external world, e.g. look at the program → effect

#### 3. Source

- **Implicit** (changing conditions according to self-updates) vs. **explicit** (**internal** or **external** occurred/computed/detected events) (e.g. production rules vs. ECA rules)
- **By request** (query on database/knowledge base or call to external system) vs. **by trigger** (e.g. incoming event message, publish-subscribe, agent protocol / coordination)
- **Internal database/KB update events** (e.g. add, remove, update, retrieve) or **external explicit events** (inbound event messages, events detected by external systems): **belief update and revision**
- **Generated/Produced** (e.g. phenomenon, derived action effects) vs. **occurred** (detected or received event)

### II) Classification of the Action Space

**Similar dimensions as for events (see above)**

**1. Temporal KR event/action perspective**:
(e.g. Event, Situation, Fluent Calculus, TAL)

- Actions with effects on changeable properties / states, i.e. actions ~ events
- <u>Focus</u>: reasoning on effects of events/actions on knowledge states and properties

**2. KR transaction, update, transition and (state) processing perspective**:
(e.g. transaction logics, dynamic LPs, LP update logics, transition logics, process algebra formalism)

- Internal knowledge self-updates of extensional KB (facts / data) and intensional KB (rules)
- External actions on external systems via (procedural) calls, outbound messages, triggering/effecting
- Transactional updates possibly safeguarded by post-conditional integrity constraints / test case tests
- Complex actions (sequences of actions) modeled by action algebras (~event algebras), e.g. delayed reactions, sequences of bulk updates, concurrent actions
- <u>Focus</u>: declarative semantics for internal transactional knowledge self-update sequences (dynamic programs)

**3. Event Messaging / Notification System perspective**
- Event/action messages (inbound / outbound messages)
- Often: agent / automated web) service communication; sometimes with broker, distributed environment, language primitives (e.g. FIPA ACL) and protocols; event notification systems, publish / subscribe
- <u>Focus</u>: often follow some protocol (negotiation and coordination protocols such as contract net) or publish-subscribe mechanism

**4. Production rules perspective:**
(e.g. OPS5, Clips, Jess, JBoss Rules/Drools, Fair Isaac Blaze Advisor, ILog Rules, CA Aion, Haley, ESI Logist )

- Mostly forward-directed non-deterministic operational semantics for Condition-Action rules
- Primitive update actions (assert, retract); update actions (interpreted as implicit events) lead to changing conditions which trigger further actions, leading to sequences of triggering production rules
- But: approaches to integrate negation-as-failure and declarative semantics exist, e.g. for subclasses of production rules systems such as stratified production rules with priority assignments or transformation of the PR program into a normal LP
- Related to serial Horn Rule Programs

**5. Active Database perspective**:
(e.g. ACCOOD, Chimera, ADL, COMPOSE, NAOS, Hi-Pac)

- ECA paradigm: "*on Event and Condition do Action*"; mostly operational semantics
- Instantaneous, transient events and actions detected according to their detection time
- Complex events: event algebra (e.g. Snoop, SAMOS, COMPOSE) and active rules (sequences of self-triggering ECA rules)

<u>III) Classification of the Event / Action / State Processing respectively Reasoning Space</u>

**1. Event/Action Definition Phase**
- Definition of event/action pattern by event algebra
- Based on declarative formalization or procedural implementation
- Defined over an atomic instant or an interval of time, events/actions, situation, transition etc.

**2. Event/Action Selection Phase**
- Defines selection function to select one event from several occurred events (stored in an event instance sequence e.g. in memory, database/KB) of a particular type, e.g. "*first*", "*last*"
- Crucial for the outcome of a reaction rule, since the events may contain different (context) information, e.g. different message payloads or sensing information
- **KR view**: Derivation over event/action history of happened or future planned events/actions

**3. Event/Action Consumption / Execution Phase**
- Defines which events are consumed after the detection of a complex event
- An event may contribute to the detection of several complex events, if it is not consumed
- Distinction in event messaging between "multiple receive" and "single receive"
- Events which can no longer contribute, e.g. are outdated, should be removed
- **KR view**: events/actions are not consumed but persist in the fact base

**4. State / Transition Processing**
- Actions might have an internal effect i.e. change the knowledge state leading to state transition from (pre)-condition state to post-condition state.
- The effect might be hypothetical (e.g. a hypothetical state via a computation) or persistent (update of the knowledge base),
- Actions might have an external side effect

➔ Separation of this phases is crucial for the outcome of a reaction rule based system since typically event occur in a context and interchange context date to the condition or action (e.g. via variables, data fields).

Declarative configuration and semantics of different selection and consumption policies is desirably

## 4. Discussion of Relevant Work

In the domain of active database several prototypes have been developed, e.g. ACCOOD [1], Chimera [2], ADL [3], COMPOSE [4], NAOS [5], HiPac [6], AMIT [7], RuleCore[3]. These systems mainly treat complex event detection purely procedural or on a simple level and focus on specific aspects. Other papers discuss formal aspects on a more general level - see [10] for an overview.

There has been also a lot of research and development concerning knowledge updates and active rules (execution models) in the area of active databases and several techniques based on syntactic (e.g. triggering graphs [13] or activation graphs [14]) and semantics analysis (e.g. [15], [16]) of rules have been proposed to ensure termination of active rules (no cycles between rules) and confluence of update programs (always one unique outcome). The combination of deductive and active rules has been also investigated in different approaches manly based on the simulation of active rules by means of deductive rules [17-19, 59]. These approaches often assume a very simplified operational model for active rules without complex events and ECA related event processing.

Several event algebras have been developed, e.g. Snoop [11], SAMOS [12], ODE [4]: The object database ODE [4] implements event-detection mechanism using finite state automata. SAMOS [12] combines active and object-oriented features in a single framework using colored Petri nets. Associated with primitive event types are a number of parameter-value pairs by which events of that kind are detected. SAMOS does not allow simultaneous atomic events. Snoop [11] is an event specification language which defines different restriction policies that can be applied to the operators of the algebra. Complex events are strictly ordered and cannot occur simultaneously. The detection mechanism is based on trees corresponding to the event expressions, where primitive event occurrences are inserted at the leaves and propagated upwards in the trees as they cause more complex events to occur. A feature that is common to all these event algebras is that event operators are the only way to specify the semantics of complex events and as we have shown in this paper due to the treatment of complex events as occurrences at a single point in time, rather than over an interval, some operators show several irregularities and inconsistencies.

---
[3] ruleCore: http://www.rulecore.com

In the area of knowledge representation events have been also studied in different areas. Several Datalog extensions such as the LDL language of Naqvi and Krishnamurthy [52] which extends Datalog with update operators including an operational semantics for bulk updates or the family of update language of Abiteboul and Vinau [53] which has a Datalog-style have been proposed. A number of further works on adding a notion of state to Datalog programs where updates modelled as state transitions close to situation calculus has been taken place [18, 51]. The situation calculus [60] is a methodology for specifying the effects of elementary actions in first-order logic. Reiter further extended this calculus with an induction axiom specified in second-order logic for reasoning about action sequences. Applying, this Reiter has developed a logical theory of database evolution [61], where a database state is identified with a sequence of actions. This might be comparable to the Event Calculus, however, unlike the EC the situation calculus has no notion of time which in many event-driven systems is crucial, e.g. to define timestamps, deadline principles, temporal constraints, or time-based context definitions. Transaction Logics [27] is a general logic of state changes that accounts for database updates and transactions supporting order of update operations, transaction abort and rollback, savepoints and dynamic constraints. It provides a logical language for programming transactions, for updating database views and for specifying active rules, also including means for procedural knowledge and object-oriented method execution with side effects. It is an extension of classical predicate calculus and comes with its own proof theory and model theory. The proof procedure executes logic programs and updates to databases and generates query answers, all as a result of proving theorems. Although it has a rich expressiveness in particular for combining elementary actions into complex transactional ones it primarily deals with database updates and transactions but not with general event processing functionalities in style of ECA rules such as situation detection, context testing in terms of state awareness or external inputs/outputs in terms of event/action notifications or external calls with side effects.

Multi-dimensional dynamic logic programming [54], LUPS [55], EPI [56], Kabul [57] Evolp [55], ECA-LP [62] are all update languages which enable intensional updates to rules and define a declarative and operational semantics of sequences of dynamic LPs. While LUPS and EPI support updates to derivation rules, Evolp and Kabul are concerned with updates of reaction rules. However, to the best of our knowledge only ECA-LP [31] supports a tight homogenous combination of reaction rules and derivation rules enabling transactional updates to both rule types with post-conditional (ECAP) verification, validation, integrity tests (V&V&I) on the evolved knowledge state after the update action(s). Moreover, none of the cited update languages except of ECA-LP supports complex (update) actions and events as well as external events/actions. Moreover these update languages are only concerned on the declarative evolution of a knowledge base with updates, whereas the Event Calculus as used in ECA-RuleML/ECA-LP is developed to reason about the effects of actions and events on properties of the knowledge systems, which hold over an interval.

Concerning mark-up languages for reactive rules several proposals for update languages exist, e.g. XPath-Log [63], XUpdate [64], XML-RL [65] or an extension to XQuery proposed by Tatarinov et.al. [66]. This has been further extended with general event-driven and active functionalities for processing and reasoning about arbitrary events occurring in the Web and other distributed systems as well as triggering actions. Most of the proposals for an ECA or reactive web language are intended to operate on local XML or RDF databases and are basically simply trigger approaches, e.g. Active XML [67], Active Rules for XML [68], Active XQuery [69] or the ECA language for XML proposed by Bailey et.al [70] as well as RDFTL for RDF [71]. XChange [72] is a high level language for programming reactive behaviour and distributed applications on the Web. It allows propagation of changes on the Web (change) and event-based communication (reactivity) between Web-sites (exchange). It is a pattern-based language closely related to the web query language Xcerpt with an operational semantics for ECA rules of the form: "*Event query → Web query → Action*". Events are represented as XML instances that can be communicated and queried via XChange event messages. Composite events are supported, using event queries based on the Xcerpt query language [73] extended by operators similar to an event algebra applied in a tree-based approach for event detection.

## 5. Conclusion

**Reaction RuleML** is a general, practical, compact and user-friendly XML-serialized language for the family of reaction rules. It incorporates different kinds of production, action, reaction, and KR temporal/event/action logic rules into the native RuleML syntax using a system of step-wise extensions. In particular, the approach covers different kinds of reaction rules from various domains such as active-database ECA rules and triggers, forward-directed production rules, backward-reasoning temporal-KR event/action/process logics, event notification & messaging and active update, transition and transaction logics. It covers different **execution styles** for processing them such as **active**, where the reaction rules actively pull or detect the events possibly clocked by a monitoring/validity time function, **passive**, where the reaction rules passively wait (listen) on matching event instances, e.g. incoming event messages, which match with the define event definition patterns, and **reasoning**, where the focus is on the formalization of events and actions and reasoning on their effects on changeable knowledge states (fluents). Reaction rules can be specified on a global level in a tight combination with other rule types such as derivation rules or integrity constraints or locally, i.e. nested within other derivation or reaction rules. There are different **evaluation styles** for reaction rules such as strong and weak interpretation which are used to manage the "justification lifecycle" of local reaction rules in the derivation process of the outer rules. Reaction RuleML supports **procedural calls** on external procedures with side effects and enables expressive **transactional OID-based updates** on the extensional and intensional knowledge base, i.e. on facts and rules. Sophisticated postconditional verification, validation and integrity tests (V&V&I) using integrity constraints or test cases can be applied as post-conditional tests, which possibly might lead to roll-backs of the update actions. Complex events and actions can be expressed in terms of **complex event / action algebra operators** and different selection and consumption policies can be configured.

In this white paper we have defined important constructs and classified the event/action/state definition and processing/reasoning space of Reaction RuleML.

**Note:** The contents of this white paper and the classification scheme has been extracted from the ECA-LP / ECA-RuleML work - see e.g. technical report:

Paschke, A.: ECA-RuleML: An Approach combining ECA Rules with temporal interval-based KR Event/Action Logics and Transactional Update Logics, Internet-based Information Systems, Technical University Munich, Technical Report 11 / 2005.